# The Polish Vocabulary Size Test: A Novel Adaptive Test for Receptive Vocabulary Assessment

Danil Fokin [1], Monika Płużyczka [1], Grigory Golovin [2]

1 University of Warsaw, Poland, d.fokin@uw.edu.pl, https://orcid.org/0000-0003-0667-835X

1 University of Warsaw, Poland, https://orcid.org/0000-0003-2013-8261

2 Independent, USA, https://orcid.org/0000-0003-1636-8507

# Introduction

Vocabulary size is a crucial component of language proficiency (Laufer & Nation, 2001; Qian & Schedl, 2014), enhancing reading comprehension, -proficiency, and -efficiency (Engku et al., 2016; Chateau & Jared, 2000; Masrai, 2019; Schmitt, 2014; Tschirner, 2021), as well as speed of word and meaning recognition (Allal-Sumoto et al., 2024; Laufer & Nation, 2001; Lemhöfer et al., 2008). However, vocabulary is not a homogeneous continuum, being categorised into receptive and productive types (Laufer & Goldstein, 2004; Laufer & Paribakht, 1998; Meara, 1990; Webb, 2005). Since receptive vocabulary is acquired by reading and listening and refers to form-to-meaning mapping (Laufer & Paribakht, 1998; Schmitt, 2014), most vocabulary size tests are aimed at assessing the passive part of the lexicon (Stoeckel et al., 2021; Webb, 2021).

Both receptive and productive vocabularies are modulated by many factors, e.g. education, time spent reading/listening, or multilingualism (Guash et al., 2023; Mainz et al., 2017; Kuperman & van Dyke, 2013; Vermeiren et al., 2023). Age of acquisition has shown a consistent correlation with vocabulary size among native speakers in Dutch (Keuleers et al., 2015), English (Brysbaert et al., 2016), German (Boone & De Wilde, 2023), and Catalan (Guash et al., 2023).

The effective vocabulary test is expected to distinguish natives and non-natives based on vocabulary breadth, and establish a correlation between vocabulary size and age for native speakers. In the paper, we are presenting the pilot results of the adaptive Polish Vocabulary Size Test (PVST) which ought to become a reliable tool for testing receptive vocabulary of native and non-native Polish speakers.

## Literature Review

### Vocabulary Size Tests

There are many ways to measure vocabulary size: paper-and-pencil tasks (e.g Kavé & Halamish, 2015), subscales of Wechsler's IQ test (Hartshore & Germine, 2015; Sohacka, 2016) and Nelson-Denny Reading test (Andrews et al., 2020; Vermeiren et al., 2023), Peabody Picture Vocabulary Test (Bohn et al., 2024; Carroll et al., 2016; Mainz et al., 2017), or Vocabulary Levels Test (Schmitt et al., 2001). Nowadays, the two most popular ones are the Vocabulary Size Test (VST) (Nation & Beglar, 2007) and LexTale (Lemhöfer & Broersma, 2012).

VST has versions to examine receptive or productive vocabulary (Nation, 2024), and adapted to Japanese (Derrah & Rowe, 2015), Korean (Park, 2024), Mandarin (Zhao & Ji, 2018), and Persian (Karami et al., 2012). In the VST, items are clustered based on frequencies and participants select either a synonym or word definition out of the list of items. LexTale was translated into German, Dutch, and English (Lemhöfer & Broersma, 2012), French (Brysbaert, 2013), Italian (Amenta et al., 2021), Spanish (Ferré & Brysbaert, 2017), Portuguese (Zhou & Li, 2022), and Chinese (Chan & Chang, 2018; Qi et al., 2022). It is based on the lexical decision task where participants decide whether the string of letters is a real word.

Both tests are mentioned as having limitations. Coxhead et al. (2014) argued that guessing in the multiple-choice task (main VST principle) may inflate scores in the meaning recognition task and overestimate vocabulary knowledge. Stoeckel et al. (2021) mentioned the scoring interpretation, number and types of items are critical shortcomings of the tests that lead to vocabulary size overestimation (for further discussion see Webb, 2021; Read, 2023). LexTale was criticised for being overestimated in its reliability while the original study of Lemhöfer & Broersma (2012) lacks replications; it is not well-adapted to capture differences in second language proficiency (Puig-Mayenco et al., 2023).

The alternative to the above-mentioned methods is Computerised Adaptive Testing (CAT) which is based on Item Response Theory (IRT). The popularity of this method is rising (Bohn et al., 2024; Chan & Chang, 2018; Gibson & Steward, 2014; Tseng, 2016). The main advantage of this approach is the automatic selection of the upcoming stimulus based on the participant's previous response which increases reliability and precision while minimising time- and cognitive-consumption. Based on this methodology, Golovin (2015) developed an Adaptive online Vocabulary Size Test (AoVST) for the Russian language. AoVST demonstrated strong nonlinear relations between receptive vocabulary and age and high validity based on more than 400 thousand responses. AoVST is adapted to English, German,

Ukrainian, Tatar, and Hebrew and has been used in several psycholinguistic studies (Maslennikova et al., 2017; Parshina et al., 2024; Westergaard, 2024). A similar approach is employed in the present study.

**Vocabulary Testing in Polish**

For Polish, the number of tools to examine one's vocabulary size is limited. One of the methods used is a verbal subscale of Wechsler's test for adolescents (WAIS-R). However, it is only a part of a cognitive battery, being not designed to measure vocabulary exclusively. Sochacka (2019) indicated WAIS-R volatility due to the Flynn effect, i.e. "the gradual cross-cultural rise in raw scores obtained on measures of general intelligence" (American Psychological Association, 2018) and mentioned that test takers of the children's version (WISC-R) outperform adolescence in some subscales (Sohacka, 2019). Thus, the Polish adaptation of Wechsler's test is outdated and its reliability is understudied.

Recently, Muszyński et al. (2023) introduced an adaptation of the Pathfinder General Cognitive Ability Test (PGCAT) which contains verbal and vocabulary subscales, including the unpublished version of LexTale-PL. PGCAT demonstrated high validity and preciseness in intelligence testing. However, the vocabulary subscale is unlikely applicable for size testing. Researchers do not separate productive and receptive vocabularies and employed a non-standardized approach for stimulus and synonym selection, including direct translations from English, equivalent substitutions, brainstorming sessions, and reliance on frequencies extracted from Duolingo. This lack of transparency potentially undermines the validity of the stimulus selection method, making the vocabulary subscale not sensitive to vocabulary size differences.

**The current study**

The goal of the present study is to bridge the gap in Polish vocabulary testing by developing a tool that is both reliable and requires little time and cognitive effort. We developed an adaptive Polish Vocabulary Size Test (PVST) which is focused on testing receptive vocabulary and language proficiency rather than cognitive constructs, making it relevant for linguistic research. Using adaptive testing, we addressed the drawbacks of other tests. To validate the test, we formulated the following research questions:

1. Is the item pool used in PVST sufficiently representative for evaluating the receptive vocabulary size of Polish language users?

2. Does the methodology used effectively capture underlying cognitive processes associated with vocabulary knowledge?

3. Is PVST reliable across different age groups, native and non-native speakers?

## Test design

The Polish Vocabulary Size Test (PVST) is based on Item Response Theory (IRT) (Reise & Waller, 2009; Embretson & Reise, 2000) and Computerized Adaptive Testing (CAT) (Straetmans & Eggen, 1998; Eggen, 2018). It uses binary (yes/no), multiple-choice, and pseudoword stimuli (see *Stimuli Types* and *Stimuli Selection* sections for more details). A stimulus is presented and assesses whether the test-taker knows that word. Based on the responses, we estimate the test-taker's ability and its associated uncertainty using standard IRT procedures (see the *Scoring* section for details). We then followed a CAT approach, selecting the next test item to match its difficulty to the test-taker's estimated ability level. The process continues until 30 stimuli have been presented. We conducted four pilot tests and until we found the number of stimuli which strikes a good balance between the duration of the test (about 2 minutes) and its precision and accuracy (see the Results – *Rasch Analysis section* for details). Finally, we convert the estimated ability from logits (IRT units) to words, making the results more meaningful for test-takers. We also compute an attention index, which serves as a measure of result trustworthiness (see the *Scoring* section for further details).

Computerized Adaptive Testing, which is central to the PVST, has proven to be a powerful approach to vocabulary assessment, offering efficiency, precision, and a customized experience unmatched by traditional tests. Across diverse populations, CAT-based vocabulary tests deliver shorter yet equally reliable measures of vocabulary size (Mizumoto et al., 2017; Tseng, 2016). They achieve this through intelligent item selection tailored to each individual's proficiency level, resulting in precise scores while minimizing unnecessary testing. Indeed, presenting simple stimuli to proficient native speakers or difficult stimuli to beginner learners would provide little useful information. By matching each stimulus to the current estimate of the test-taker's ability, we maximize the amount of information each stimulus contributes to the test. Furthermore, the individualized nature of CAT broadens accessibility (one test fits all proficiency levels) and enhances the test-taking experience, potentially reducing anxiety related to overly easy or difficult stimuli and boosting participant engagement (Kachergis et al., 2022; Ling et al., 2017)

### Stimuli types

The test uses binary, multiple-choice, and pseudowords stimuli. For binary stimuli, a test taker chooses whether he/she knows or does not know that stimulus (self-assessment). These items are low-demanding, easy to create, and efficient in testing large populations (Harrington & Carey, 2009; Pellicer-Sánchez & Schmitt, 2012). Many such items can be presented in a relatively short time frame, achieving high test precision without putting too much load on test takers. However, binary stimuli have limitations (see Beeckmans et al., 2001; Mochida & Harrington, 2006). They include *overconfidence bias* (American Psychological Association, 2018), i.e. participants decide by themselves whether they know the meaning, *self-report bias*, i.e. participants' desire to appear more knowledgeable and meet expectations (Bauhoff, 2014), and *high false alarm rate* (Mochida & Harrington, 2006).

For multiple-choice stimuli, a test taker also initially chooses whether he/she knows or does not know that stimulus. If the "I know this word" option is selected, we present four options – a synonym and three distractors – and ask that the test taker clarify the meaning of the stimulus.

Pseudowords are presented similar to binary and multiple-choice stimuli. If a test taker indicates that he/she knows that word, we show a warning, but do not penalize the result.

The rationale for using three stimulus types is to capitalize on their complementary strengths and offset their individual limitations. Binary real-word stimuli allow for rapid testing with numerous items but suffer from high guessing probability, inflated false alarm rates, and self-report bias. Multiple-choice items significantly reduce guessing and enhance test reliability by providing more discriminative power among test-takers, yet they increase cognitive load and test duration. Pseudowords further minimize guessing probability and enhance item variability. Together, these stimulus types create a balanced and efficient assessment tool, maximizing reliability and validity while maintaining brevity in testing.

Currently, there is no consensus regarding the optimal proportion of real words (providing information on respondent's vocabulary) to pseudowords (providing control of respondent's attention and accuracy) in language tests (Beeckmans et al., 2001). Keuleers et al. (2015) used a ratio of 75% real words to 25% pseudowords, justifying this choice by the opportunity to "collect more responses to words and therefore more data." They argued that using equal numbers of real and pseudowords would result in participants knowing fewer than 50% of the items (Keuleers et al., 2015, p. 1670). Lemhöfer et al. (2008) also adopted a similar proportion in their bilingual test employing a lexical decision task. Conversely, Mainz et al. (2017) used equal numbers of real and pseudowords in their lexical decision task, motivated by the need to compare reaction times and decision-making processes between these two item types. Pellicer-Sanchez and Schmitt (2012) considered 30% pseudowords a suitable compromise for their yes/no vocabulary test.

For the current study, we chose a proportion of 60-20-20 for binary, multiple-choice, and pseudowords stimuli respectively, since it strikes a good balance between information (coming from binary and multiple-choice stimuli) and control (coming from multiple-choice and pseudoword stimuli).

### Stimuli selection

We applied the following criteria for stimuli selection:

1. Binary and multiple-choice stimuli span from high- to low-frequent words, aiming to cover all language levels from beginners to native speakers. These words address general topics, do not have domain specification, and primarily represent modern Polish language. We tried to ensure that selected items represent unique Polish lexemes to minimise positive interference from another language by appealing to Polish dictionaries. Exceptions are made only for the most difficult, low-frequent items, e.g. *burłak* (burlak), *żyrandol* (girandole).

2. The synonyms used as multiple-choice options are hyponyms, equivalents, or close collocates of the stimuli. These synonyms have higher frequencies than the stimuli, so if test-takers know a stimulus, they most likely know its synonym as well. In some exceptions, synonyms are presented as two-word phrases.

2a. Distractors in multiple-choice stimuli are semantically distant from the stimuli and represent the same part of speech, reducing the guessing probability.

3. Pseudowords were randomly chosen from Imbir et al. (2015) list, being judged by participants as good examples of pseudowords.

Initially we chose stimuli based on frequency provided in plTenTen19 web corpus (Lexical Computing CZ, n.d.; Jakubíček et al., 2013). Once data is collected, the test is going to self-calibrate based on responses, not frequencies. It is a popular index for selecting stimuli (see VST and its adaptations (Nation, 2024)). Although some studies showed that not the frequency alone but print exposure, education, source corpus, and lexical features impact vocabulary size estimation (Hashimoto, 2021; Kuperman & van Dyke, 2013). However, Schmitt (2014) noted that with the decrease in frequency the gap between receptive and productive vocabularies increases and Monaghan et al. (2017) argued that print exposure is the main contributor to variation in both frequency and vocabulary learning. In the present study, frequency is used for initial stimuli selection.

To extract stimuli we applied plTenTen19 Polish web corpus in SketchEngine (Lexical Computing CZ, n.d.; Jakubíček et al., 2013). Along with a huge volume (> 18 million unique words and lemmas, > 13 M. documents), it contains a wide range of sources, including websites, titles, documents, wiki pages, etc. providing recent collocations and contextual usage. For the initial item pool we also used the age of acquisition as mentioned in Imbir et al. (2016). All selected items were further cross-checked using Polish dictionaries (PWN, https://sjp.pwn.pl/; WSJK, https://wsjp.pl/), and the national corpus (NKJP, https://nkjp.pl/).

The stimuli list was compiled in a stepwise fashion: 1) initial test-retest compilation that includes inspection by the native language linguist; 2) pre-test on the small sample (up to 50 test takers); 3) piloting of the item pool at the website. All test versions and revisions are summarised in Table 1. The pre-final (3rd) set was published on the website myvocab.info/pl, authors' personal Facebook pages, and Reddit platform to collect representative dataset.

**Table 1**

*The changes made in the stimuli set in the course of the pilot study*

| Set version | Total stimuli (incl. pseudowords) | Binary questions | Multiple-choice questions | Pseudowords |
|---|---|---|---|---|
| 1st | 233 | 146 | 49 | 38 |
| 2nd | 290 | 176 | 76 | 38 |
| 3rd | 291 | 181 | 72 | 38 |
| 4rd | 233 | 140 | 55 | 38 |

**Scoring**

The results of the test consist of two estimates: test-takers' ability (latent variable), i.e. level of a trait (see Nevin et al., 2015), and attention index (a measure of trustworthiness of the result). We obtained an estimate of test takers' ability in logit units, which is a standard for any IRT-based test. Although these units are optimal for psychometric research, they are not easy to understand for the general public. We therefore converted it from logits to words using the logit function: $y=a/[1+exp(-b*(x-c))]$, where $x$ – test takers' ability in logits, $y$ – vocabulary size in words, and $a, b, c$ – conversion coefficients. We obtained the coefficients from fitting stimuli rank (their order from high to low frequency based on plTenTen19 corpus data) vs stimuli difficulty (estimated from the test) using the same function.

The intuition for this conversion is the following. If we take a stimulus with a difficulty equal to a test-taker's ability, the probability that the test-taker knows that word is 50%. If we order all words from most common to least common, that stimulus (which ended up with number N in the list), would split all words into two parts. The more common part (with word numbers <$N$) should contain some words which the test taker does not know. However, the less common part (with word numbers >$N$) should also contain some words, which the test taker does know. These two groups of words cancel each other. As a result, it is possible to say that the test taker knows approximately $N$ words. Following one of the largest Polish dictionaries – the PWN orthographic dictionary (https://shorturl.at/0Qqfb) – we assumed the total number of Polish words to be 140.000. Thus, we set coefficient $a$ to 140.000, bounding the estimated vocabulary size between 0 and 140.000 words.

Attention index, which measures the trustworthiness of the results, was calculated using a formula (x+y)/(ax+ay), where x represents the number of pseudowords marked as unknown, ax is the total number of presented pseudowords, y is the number of multiple-choice test words defined correctly, and ay is the total number of presented multiple-choice questions. Attention index reaches 100% when a test taker responds "I don't know" to all pseudowords, and chooses correct definitions of all multiple-choice test words in the test. For individual tests, we did not use the attention index to correct (penalise) the estimation of vocabulary size. We considered all results with an attention index below 70% as not trustworthy and did not include them in the estimation of test word difficulties. We also filtered such results out when we aggregated test results.

### Validation study

#### Data Analysis

The data underwent pre-processing, where missing values and outliers were removed using a standard cleaning procedure (Mean ± 2SD). We filtered out the results of test takers who guessed too much (attention index below 70%) and whose test duration was less than 60 seconds. Due to the outsourced nature of our data collection, we did not have full control over participants' self-reported demographic information, including age and language nativeness. This occasionally resulted in implausible entries, such as vocabulary scores inconsistent with known acquisition patterns or improbable ages (e.g., 0 or 100 years). To address this, we applied a data-cleaning procedure in two steps. First, we excluded participants whose reported age was below 7 years, considering ages below 7 as unreliable for consistent self-report or valid test performance. Second, to filter likely invalid nativeness data, applying a standard outlier cleaning procedure based on the Mean ± 2SD rule within each group (native vs. non-native). It allowed us to identify and exclude cases where the vocabulary score was either unreasonably high for self-identified non-native speakers or unreasonably low for native speakers. After cleaning, the maximum score among non-native participants was 23.394, while the minimum score for native speakers was 19.556, which we consider plausible within the test's difficulty range and scoring structure.

Specifically, we counted responses occurring within 5 minutes of each other (as more than 95% of test attempts were completed within this time), for which the reported age and native/learner status matched. We estimated that up to 10% of the data might have come from respondents who took the test multiple times. We did not exclude these responses from the analysis; however, we believe this issue does not significantly affect the results.

Variables were visually and statistically assessed for normality. Shapiro-Wilk's test was conducted to assess the normality of the distribution of key variables. Pearson's correlation was applied to explore relations between three key variables: vocabulary size, age, and attention index. To establish the fitness of the stimuli selected, the Rasch model was built to calculate infit outfit statistics for each stimulus and participant, and to reveal how well the test predicts the participant's ability. Finally, an independent sample t-test was performed to observe differences in vocabulary size between native and non-native Polish speakers.

**Participants**

The participants were volunteered via Riddle platform, Facebook, and personal invitations. After applying a rigorous data-cleaning pipeline, we retained 656 high-quality observations out of an initial 1475 participants (Table 2). Specifically, we removed participants who scored less than 75 on attention index (meaning that they marked too many pseudowords as known and chose too many wrong definitions of multiple-choice stimuli), and those who took too little (<30 seconds) time for the test. We chose to keep the results of 4% of respondents who took a significant time (>6 minutes, or >12 seconds per stimulus) doing the test, since additional analysis showed that their results do not differ from the rest. This filtering process, though substantial, ensured quality of our data for further statistical analysis. The native group is larger (417 vs 239) and around eight years older (see Table 4 below) than the non-native one.

**Table 2**

*Sample size and characteristics of norm sample and two subsamples: Polish and non-Polish speakers*

|  | | Samples | |
|---|---|---|---|
|  | All | Polish | non-Polish |
| N | 656 | 417 | 239 |
| M_age (SD) | 30.5 (11.7) | 33.4 (12.2) | 25.3 (8.6) |
| Age range | 8–66 | 12–66 | 8–49 |

**Procedure**

Participants took the test either online from personal computers or smartphones. The study had no time limits (M<sub>test_dutation</sub> ≈ 2.5 min). First, participants were instructed that they would see real and pseudowords and their task was to decide whether they knew the item. We specified "knowing" as "the ability to provide at least one of the word meanings". Then, the participant responded to the question "Do you know the word on the screen?" ("Czy Państwo znają słowo na ekranie?") and chose between two options: "I know" ("znam") and "*I don't know*" ("nie znam").

Multiple-choice questions were linked to particular items from the pool and presented similarly to binary ones. If the participant responded "I don't *know* ("nie znam") to the binary question, the stimulus was marked as "unknown" and the test continued. But if the test taker responded "I know" to the multiple-choice item, s/he was further obliged to "*define the word meaning*" ("określ znaczenie słowa") by selecting one out of four items. All items were randomised.

After completing the test, participants were requested to provide their age and indicate whether their native language was Polish or non-Polish. They were also asked to confirm if they answered honestly; responses from those who did not check this box were not counted. Following this, participants received their results, which included an estimate of their receptive vocabulary size in words and statistics comparing native and non-native Polish speakers.

## Results

### Rasch analysis

To evaluate precision of PVST, quality of stimuli, and participants' performance, we ran a dichotomous Rasch model using the TAM package in R (Robitzsch et al., 2020) and following the guideline of Wind & Hua (2021). The Rasch equation serves to validate the psychometric test, calculating an interval-level estimation of a person's ability, i.e. accuracy of test performance, and items difficulty on a linear scale (Reise & Waller, 2009; Wind & Hua, 2021). This method has been already used for vocabulary size test validations (Akase, 2021; Beglar, 2010; Runnels, 2012).

Table 3 summarises logit-scale calibrations, SDs, and data fitness based on 1056 observations of 252 stimuli items. Average values of model fit statistics indicate a high degree of reliability with outfit and infit mean square (MSE Mean) statistics around 1.00, and standardised outfit and infit values close to 0.00. High values of the reliability of separation, which is similar to Cronbach's alpha, demonstrate PVST high precision in separation of test takers based on their language proficiency, and confirm that the latent trait estimates accurately reflect true individual differences (in both stimuli and respondents) with minimal measurement error. PVST demonstrated excellent precision with a STRATA value of 9.5, indicating the capacity to reliably differentiate respondents into approximately nine statistically meaningful levels.

**Table 3**

*The Rasch model summary*

| Statistics | Items (SD) | Person (SD) |
|---|---|---|
| Logit Scale Location Mean | -1.51 (2.66) | .18 (3.12) |
| **Outfit MSE Mean** | .81 (.25) | .71 (.37) |
| **Infit MSE Mean** | .91 (.14) | .83 (.29) |
| **Std. Outfit Mean** | -1.35 (1.99) | -.14 (1.15) |
| **Std. Infit Mean** | -.54 (1.05) | -.58 (.93) |
| **Reliability of Separation** | .96 | .95 |

*Note*. MSE - Mean Squared Error; Std. - standardised

To assess the quality of individual test items, we plotted participants' response curves (Figure 1). We found that outfit and infit metrics well described the quality of the fit, i.e. how close the measured probabilities of test takers to know a test item to the expected ones. For items with outfits and infits close to 1, the item response curves showed a very good fit (Figure 1, left panel) while high infit and outfit reveal deviation from the measured probability (Figure 1, right panel). After the Rasch analysis, we excluded items with infit/outfit values exceeding 1.3 and z-score > 2.0 (Nevin et al., 2015; Wind & Hua, 2021), leaving 195 words (with addition of 38 pseudowords).

**Figure 1**

*Item response curves for a high-quality ("animusz") and low-quality ("diametralny") test item*

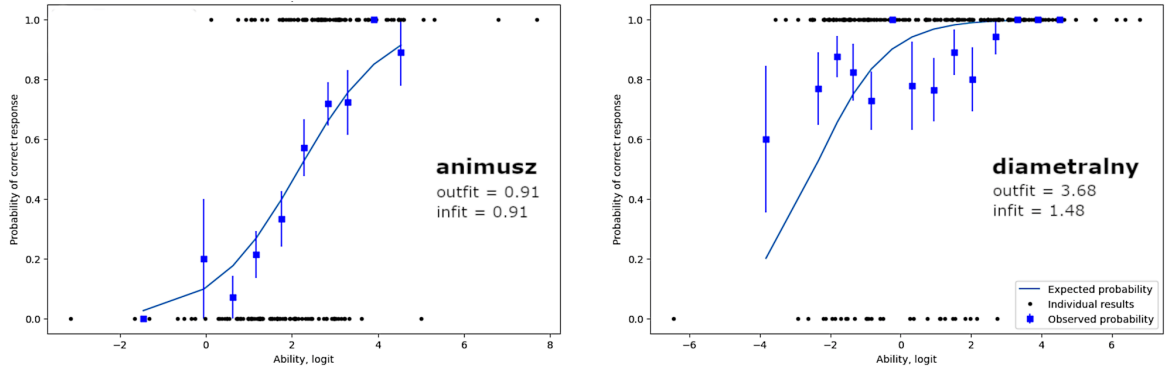

To visualize the quality of the stimuli bank, we can plot item-person (or Wright) map (see Figure 2). It shows distributions of both respondents (on the left) and stimuli (on the right), plotted along the same logit scale, which becomes ability for respondents and difficulty for stimuli.

**Figure 2**

*Item-Person Map*

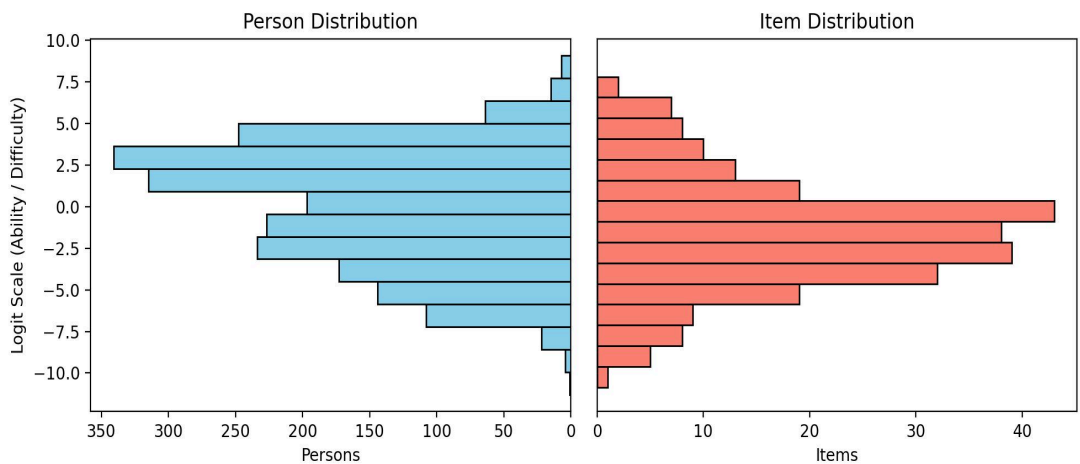

The map demonstrates that for every ability level, from beginner learners on the bottom to proficient native speakers on the top, the stimuli bank has items with corresponding levels of difficulty.

**Vocabulary Size of Polish and non-Polish speakers**

Table 4 demonstrates the results of the independent samples t-test comparing two groups: native and non-native speakers.

**Table 4**

*Between group differences in age, test duration, accuracy, and vocabulary_size*

| Variable | df | t | Effect Size Cohen's d | p | 95 % CI Cohen's d | |
|---|---|---|---|---|---|---|
| | | | | | Lower | Upper |
| Duration | 654 | − .191 | − .015 | .849 | − .175 | .144 |
| Attention index | 654 | − .503 | − .041 | .615 | − .200 | .118 |
| Age | 654 | − 9.071 | − .736 | < .001 | − .900 | − .572 |
| Vocabulary size | 654 | − 44.746 | − 3.630 | < .001 | − 3.882 | − 3.376 |

There were no differences between groups in test duration (t = −.191; d = −.015; $M_{general}$ = 113.18 sec; $SD_{general}$ = 34.21) and attention index (t = − .503; d = −.041; $M_{general}$ = 90.42; $SD_{general}$ = 8.04), suggesting both groups are equal in their test performance. The non-native group was significantly younger than the native one (t = − 9.071; d = − .736); the vocabulary size of Polish speakers was significantly larger than those of non-Polish speakers (t = − 44.746; d – 3.630).

Table 5 reflects 10 times larger vocabulary size (M = 75.125; SD = 23.055) and a wider range (range: 19.556 to 122.693) of native speakers' compared to non-speakers ones (M = 7.165; SD = 5.828; Range: 646–23.394).

**Table 5**

*Differences between Polish and non-Polish speakers in vocabulary size*

| Variable | Native Language | N | M | SD | Range | p |
|---|---|---|---|---|---|---|
| Vocabulary size | Polish | 417 | 75.125 | 23.055 | 19.556 – 122.693 | < .001 |
| | non-Polish | 239 | 7.165 | 5.828 | 646 – 23.394 | < .001 |

The distribution of native Polish speakers' vocabulary (Figure 3) is close-to-normal (skewness: − .167) while non-natives' distribution is left-skewed (skewness: .882) (Figure 3, right panel) and more dense (Figure 3, left panel).

These findings suggest that PVST is well-fitted to measure vocabulary size and distinguish speakers based on language proficiency level in a relatively short time (up to 2 minutes) and with a low cognitive load (Mean attention index ~ 90).

**Figure 3**

*Vocabulary size distribution by native and non-native Polish speakers*

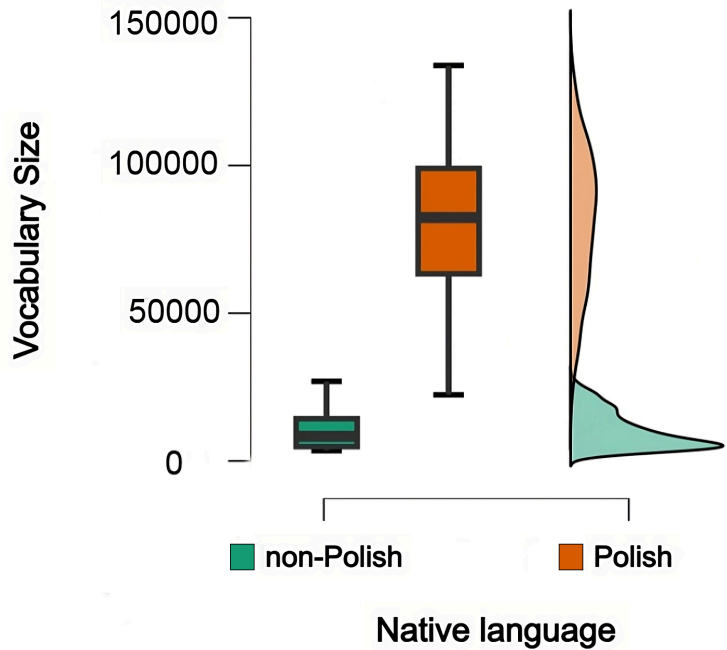

**Pearson's correlation**

Table 6 reveals strong positive correlations between vocabulary size and age (r =.496; p <.001), vocabulary size and attention index (r = .133; p <.001), and between test duration and age (r = .187; p <.001); moderately strong correlation between age and attention index (r = .085; p <.05), and strong negative link between duration and attention index (r = – .153; p <.001).

**Table 6**

*Correlations between key variables*

| Variable | 1 | 2 | 3 | 4 |
|---|---|---|---|---|
| 1. Age | — | | | |
| 2. Vocabulary_size | n = 656<br>r = .496***<br>Fisher's z = .544 | — | | |
| 3. Attention index | n = 656<br>r = .085*<br>Fisher's z = .086 | n = 656<br>r = .133***<br>Fisher's z = .134 | — | |
| 4. Duration | n = 656<br>r = .187***<br>Fisher's z = .189 | n = 656<br>r = .068<br>Fisher's z = .068 | n = 656<br>r = − .153***<br>Fisher's z = − .154 | — |

*Note. *p<.05, **p<.01, ***p<.001*

For native speakers, the correlation between the vocabulary size and age is significant (n = 417, *r* = .494, *p* < .001), while for non-native speakers it is weak (n = 239, *r* = .137; *p* < .34).

**Vocabulary size as a function of age**

Finally, to explore the impact of the native language and age on vocabulary size, we clustered participants into eleven age groups and calculated means for each group. Figure 4 demonstrates a rapid growth in vocabulary size by native speakers from 10 to 30 years old and slowing down after 30; non-Polish speakers, in turn, do not show such a tendency, having steady vocabulary size from 8 to 45 years old. The variation in vocabulary size within clusters is 2.565 to 8.566 for non-natives and 38.493 to 95.167 for native Polish speakers.

**Figure 4**

*Vocabulary size changes in native and non-native speakers by age groups*

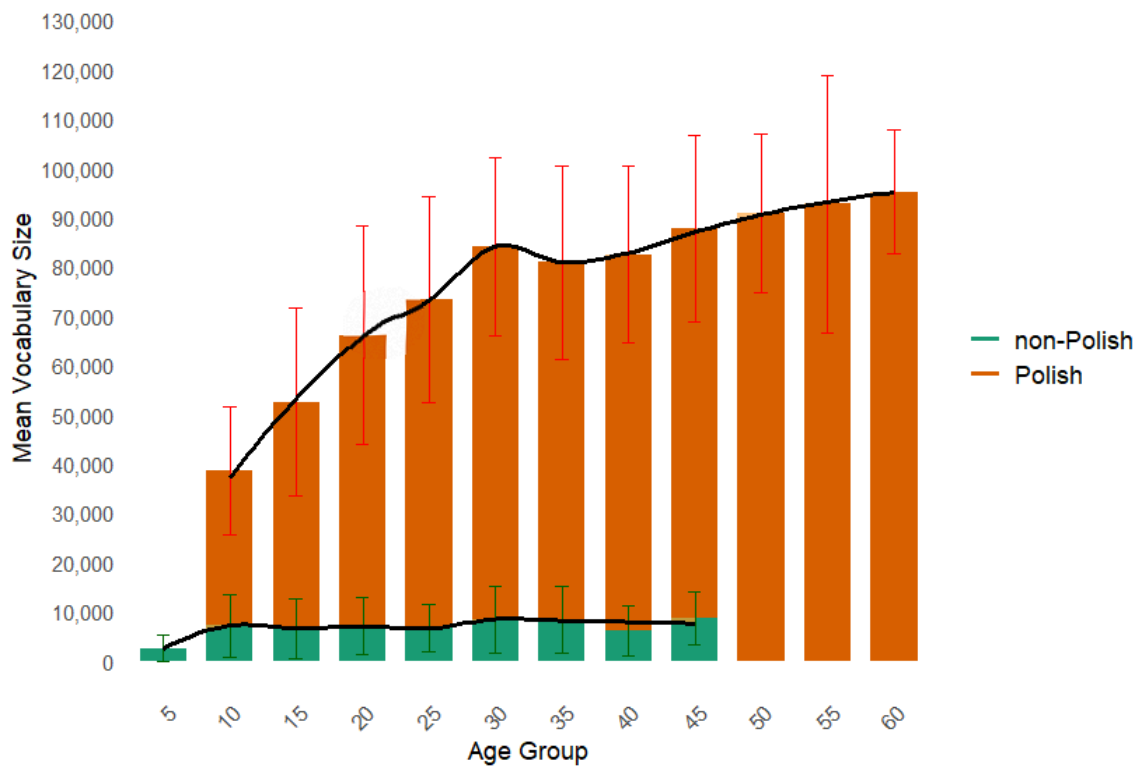

## Discussion

The main goal of the current study was to develop and pilot a novel adaptive Polish Vocabulary Size Test (PVST) using Item Response Theory and Computerized Adaptive Testing.

**RQ 1.** Is the item pool used in PVST sufficiently representative for evaluating the receptive vocabulary size of Polish language users?

To ensure PVST representativeness, the item pool includes words from various parts of speech, spanning a range of frequencies and primarily rooted in Polish, with rare exceptions for low-frequency loanwords.

PVST demonstrates strong content validity (Messick, 1988) (Table 3), confirmed through Rasch modelling and strict cut-off criteria. We excluded items with infit/outfit values exceeding 1.3 and z-score > 2.0 (Nevin et al., 2015; Wind & Hua, 2022), reducing the initial item bank from 252 to 195 stimuli (excluding pseudowords).

The final PVST set demonstrates moderate overall difficulty ($M_{xsi}$ = − 1.51; Mean $SE_{xsi}$ = .35). Item difficulty ranges from -6.85 (wieczór – evening) to 5.60 (jedlca – guard, tsuba) (see Figure 2, right panel). The item pool is skewed towards easier items, i.e. below 0, deviating from the typical range of − 3 to 3 (Embretson & Raise, 2000), revealing the imbalance of a stimulus set.

Developing a balanced item pool is inherently challenging for vocabulary size tests (Read, 2023; Stoeckel et al., 2021; Webb, 2021). Vocabulary knowledge exists on a continuum, which makes it difficult to compile a universal set of items that accurately reflects all language users' lexicons. Nevertheless, the use of adaptive testing based on IRT allows for more detailed insights into participants' word knowledge and mitigates some challenges in item selection (Embretson & Reise, 2000; Stoeckel et al., 2021).

Overall, the results of Rasch modeling align with prior research (Akase, 2022; Beglar, 2010; Bohn et al., 2024; Derrah & Rowe, 2015; Runnels, 2012), demonstrating PVST's methodological robustness and high replicability. However, while the item pool represents a range of word difficulties, its skew towards easier items may limit the test's ability to differentiate advanced vocabulary users. Future refinements should aim for a more balanced difficulty distribution to enhance representativeness across proficiency levels.

**RQ 2.** Does the methodology used effectively capture underlying cognitive processes associated with vocabulary knowledge?

Current study employed Item Response Theory (IRT) (Embertson & Reise, 2009; Reise & Waller, 2009) to develop a robust tool to assess receptive vocabulary size in Polish. High infit/outfit scores for both items and persons (Table 3) suggest that PVST targets vocabulary knowledge rather than unrelated cognitive constructs. This capability is a key principle of CAT, where the selection of upcoming items depends on participants' previous response.

PVST sensitivity to cognitive performance was further demonstrated through Pearson's correlation. Vocabulary size was positively correlated with attention index (r = .133, p < .001) while attention index is negatively connected to the test duration (r = −.153, p < .001). These findings indicate that test takers' with higher attention tend to perform better and complete the test more quickly. On average, participants with higher attention scores ($M_{attention\_index}$ = 90) completed the test within 2 minutes. Thus, PVST effectively captures the latent construct of receptive vocabulary knowledge while minimizing cognitive burden.

As expected, PVST demonstrates precision in measuring receptive vocabulary knowledge within a short duration and with minimal cognitive load. This aligns with prior findings by Embretson & Reise (2000), who showed that shorter adaptive tests could produce data comparable in accuracy to longer assessments. In this study, the high reliability of PVST, combined with its brief administration time and strong attention index, further supports its structural validity.

**RQ 3.** Is PVST reliable across different age groups, native and non-native speakers?

PVST demonstrates stability in distinguishing Polish and non-Polish speakers (Table 5), with native speakers exhibiting much larger vocabulary size compared to non-native ones (75.125 vs 7.165), aligned with prior findings (Coxhead et al., 2014; Golovin, 2015). The vocabulary sizes of non-native speakers showed a tighter distribution (SD = 5.828 words; range = 646–23.394 words) compared to the broader variability observed among native speakers (SD = 23.055 words; range = 19.556–122.693 words). This suggests greater individual differences in vocabulary breadth within the native speaker group. The PVST is also sensitive to age-related vocabulary size trends in native speakers, revealing vocabulary growth from ages 10 to 30, followed by a slight decline from 30 to 35 and then continuation of growth from 35 to 60.

These findings are consistent with those revealed in the studies in other languages (Brysbaert et al., 2016; Keuleers et al., 2015; Vermeiren et al., 2023). In contrast, no significant correlation between age and vocabulary size was observed among non-native speakers. This supports prior results indicating that vocabulary acquisition in this group is primarily driven by print exposure and practicing but not age (San Mateo-Valdehíta & Criado de Diego, 2021; Schmitt, 2014; Webb, 2005). Our findings suggest that PVST's methodology is well-suited for assessing vocabulary knowledge and reliably differentiate language proficiency among Polish language users.

In summary, the PVST effectively captures differences in vocabulary size between native and non-native speakers while demonstrating sensitivity to age-related trends in vocabulary acquisition. It aligns with established vocabulary test standards and replicates widely recognized findings. Grounded in Item Response Theory and Computerized Adaptive Testing, PVST is efficient, cognitively nondemanding, precise, and easy to administer.

The test's accessibility through an online interface further enhances its usability. Incorporating gamification elements, such as interactive features, encourages user engagement and supports test completion. These features make PVST a versatile tool with potential applications in various fields, including educational research, language instruction, psycholinguistics, and psychology.

### Limitations

The pilot study faced several limitations. **First,** remote data collection led to self-reported age and language information, resulting in approximately 50% data loss after cleaning. The study's unsupervised nature also raises the possibility of test retakes and cheating, though the remaining data were sufficient for robust analysis. **Additionally,** the stimulus list, while balanced based on Rasch analysis, may still require refinement to improve item selection. The observed vocabulary size of Polish speakers (M = 75.125) significantly differs from findings of Brysbaert et al. (2016), Guasch et al. (2023), and Keuleers et al. (2015) but align with Golovin (2015). It is likely due to statistical and lexical boundary differences: an upper bound of 140.000 words is based on the PWN dictionary.

## Ethics declarations

### Ethics Approval Statement

The study was reviewed and approved by the University of Warsaw Rector's Committee for the Ethics of Research Involving Human Participants as a part of the project "The Polish poetry through the lens of eye tracking: An expert-novice study" (Application 272/2024).

### Consent to participate

All participants gave informed consent electronically before starting the online survey. They were informed about the nature and purpose of the study, their right to withdraw at any time, and the handling of anonymized data. No personally identifying information was collected, except for age and native language.

**Consent for publication**

At the end of the survey, participants confirmed their consent for anonymized data to be used in research analysis and publication by ticking a confirmation box.

**Declaration of conflicting interests**

There are no conflicting interests to declare.

**Preregistration**

The research was not preregistered before the data collection.

**Open Practices Statement**

The study follows Open Science practices providing all results and codes provided in the open repository, available via https://osf.io/5ehjc/files/osfstorage

# Acknowledgements

We thank Polina Bakhturina ( bakhpol@gmail.com  University of Mainz) for the support and help with the project, active users of Reddit, who provided valuable comments, helping us to refine the test, and Polish linguists, who participated in the test-retest sessions, commenting and promoting the study. All these concerns were considered by the authors and helped to improve the test.

# References

Akase, M. (2022). Longitudinal measurement of growth in vocabulary size using Rasch-based test equating. *Language Testing in Asia 12*(5). https://doi.org/10.1186/s40468-022-00155-8;

Allal-Sumoto T. K.., Şahin D., & Mizuhara H. (2024). Neural activity related to productive vocabulary knowledge effects during second language comprehension, *Neuroscience Research 203,* 8-17, https://doi.org/10.1016/j.neures.2024.01.002;

Amenta S., Badan L., & Brysbaert M. (2021). LexITA: A Quick and Reliable Assessment Tool for

Italian L2 Receptive Vocabulary Size, *Applied Linguistics 42* (2), 292–314, https://doi.org/10.1093/applin/amaa020;

Andrews, S., Veldre, A., & Clarke, I. E. (2020). Measuring Lexical Quality: The Role of Spelling Ability. *Behavior research methods, 52*(6), 2257–2282. https://doi.org/10.3758/s13428-020-01387-3;

American Psychological Association. (2018). Flynn Effect. In *APA Dictionary of Psychology.* Retrieved October 23, 2024 from https://dictionary.apa.org/flynn-effect;

American Psychological Association. (2018). Overconfidence. In *APA Dictionary of Psychology.* Retrieved August 28, 2024 from https://dictionary.apa.org/overconfidence;

Bauhoff, S. (2014). Self-Report Bias in Estimating Cross-Sectional and Treatment Effects. In Michalos, A.C. (Ed.), *Encyclopedia of Quality of Life and Well-Being Research.* Springer, Dordrecht. https://doi.org/10.1007/978-94-007-0753-5_4046;

Beeckmans, R., Eyckmans, J., Janssens, V., Dufranne, M., & Van de Velde, H. (2001). Examining the Yes/No vocabulary test: some methodological issues in theory and practice. *Language Testing, 18*(3), 235-274. https://doi.org/10.1177/026553220101800301;

Beglar D. (2010). A Rasch-based validation of the Vocabulary Size Test. *Language Testing 27*(1), 101–118. https://doi.org/10.1177/0265532209340194;

Bohn M., Prein J., Koch T., Maximilian Bee R., Delikaya B., Haun D., & Gagarina N. (2024). oREV: An item response theory-based open receptive vocabulary task for 3- to 8-year-old children. *Behavioral Research 56*, 2595–2605. https://doi.org/10.3758/s13428-023-02169-3;

Boone, G., & De Wilde, V. (2023). Productive vocabulary knowledge in L2 German: Which word-related variables matter? *System*, *118*, 103150. https://doi.org/10.1016/j.system.2023.103150;

Brysbaert, M. (2013). LexTALE_FR: A fast, free, and efficient test to measure language proficiency in French. *Psychologica Belgica, 53*(1), 23–37. https://doi.org/10.5334/pb-53-1-23;

Brysbaert M., Michaël S., Paweł M., & Keuleers E. (2016). How Many Words Do We Know?

Practical Estimates of Vocabulary Size Dependent on Word Definition, the Degree of Language Input and the Participant's Age. *Frontiers in Psychology 7*. https://doi.org/10.3389/fpsyg.2016.01116 ;

Carroll R., Warzybok A., Kollmeier B., & Ruigendijk E (2016) Age-Related Differences in Lexical Access Relate to Speech Recognition in Noise. *Frontiers in Psychology* 7:990. https://doi.org/10.3389/fpsyg.2016.00990;

Chan L. & Chang B. C. (2018). LEXTALE_CH: A quick, character-based proficiency test for Mandarin Chinese. *Proceedings of the Annual Boston University Conference on Language Development 42*(1), 114 - 130.

Chateau D., & Jared D. (2000). Exposure to print and word recognition processes. *Memory and Cognition, 28*. 143–153. https://doi.org/10.3758/BF03211582;

Coxhead, A., Nation, I. & Sim, D. (2014). Creating and trialling six versions of the Vocabulary Size Test. *TESOLANZ Journal, 22*, 13-27.

Derrah, R., & Rowe, D. E. (2015). Validating the Japanese bilingual version of the Vocabulary Size Test. *International Journal of Languages, Literature and Linguistics 1*(2), 131-135;

Eggen T.J.H.M. (2018). Multi-Segment Computerized Adaptive Testing for Educational Purposes. *Frontiers in Education 11*(3), https://doi.org/10.3389/feduc.2018.00111;

Embretson S. E. & Reise P. S. (2000). *Item Response Theory for Psychologists.* Lawrence Erlbaum Associates Inc.;

Engku H., Engku I., Isarji S., & Ainon J. M. (2016). The Relationship between Vocabulary Size and Reading Comprehension of ESL Learners. *English Language Teaching, 9*(2). https://doi.org/10.5539/elt.v9n2p116;

Ferré, P., & Brysbaert, M. (2017). Can Lextale-Esp discriminate between groups of highly proficient Catalan–Spanish bilinguals with different language dominances?. *Behavioral Research 49*, 717–723. https://doi.org/10.3758/s13428-016-0728-y;

Gibson, A., & Stewart, J. (2014). Estimating learners' vocabulary size under item response theory. *Vocabulary Learning and Instruction, 3*(2), 78–84. https://doi.org/10.7820/vli.v03.2.gibson.stewart;

Golovin, G. (2015). Izmerenie passivnogo slovarnogo zapasa russkogo yazyka [Receptive vocabulary size measurement for Russian language]. *Socio-Ipsiholingvisticheskie Issledovaniyа* [Socio-Psycholinguistic Research] *3*, 148–59;

Guasch, M., Boada, R., Duñabeitia, J.A., & Ferré P. (2023). Prevalence norms for 40,777 Catalan words: An online megastudy of vocabulary size. *Behavioral Research 55*, 3198–3217. https://doi.org/10.3758/s13428-022-01959-5;

Harrington M., & Carey M. (2009). The on-line Yes/No test as a placement tool. *System 35*(4), 614-626. https://doi.org/10.1016/j.system.2009.09.006;

Hashimoto, B. J. (2021). Is Frequency Enough?: The Frequency Model in Vocabulary Size Testing. *Language Assessment Quarterly, 18*(2), 171–187. https://doi.org/10.1080/15434303.2020.1860058 ;

Imbir K., Spustek T. & Żygierewicz J. (2015). Polish Pseudo-words list: dataset of 3023 stimuli with competent judges' ratings. *Frontiers in Psychology, 6*:1395. https://doi.org/10.3389/fpsyg.2015.01395 ;

Imbir K. K. (2016). Affective Norms for 4900 Polish Words Reload (ANPW_R): Assessments for Valence, Arousal, Dominance, Origin, Significance, Concreteness, Imageability and Age of Acquisition. *Frontiers in Psychology, 7*:1081. https://doi.org/10.3389/fpsyg.2016.01081;

Jakubíček, M., Kilgarriff, A., Kovář, V., Rychlý, P., & Suchomel, V. (2013). The TenTen corpus family. *7th International Corpus Linguistics Conference CL*, 125-127.

Kachergis G., Marchman V. A., Dale PS, Mankewitz J., Frank M. C. (2022). Online Computerized Adaptive Tests of Children's Vocabulary Development in English and Mexican Spanish. *Journal of Speech Language, and Hearing Research 65*(6), 2288-2308. https://doi.org/10.1044/2022_JSLHR-21-00372

Karami, H. (2012). The Development and Validation of a Bilingual Version of the Vocabulary Size Test. *RELC Journal, 43*(1), 53-67. https://doi.org/10.1177/0033688212439359

Kavé, G., & Halamish, V. (2015). Doubly blessed: Older adults know more vocabulary and know better what they know. *Psychology and Aging*, *30*(1), 68–73. https://doi.org/10.1037/a0038669;

Keuleers, E., Stevens, M., Mandera, P., & Brysbaert, M. (2015). Word knowledge in the crowd:

Measuring vocabulary size and word prevalence in a massive online experiment. *Quarterly Journal of Experimental Psychology, 68*(8), 1665-1692. https://doi.org/10.1080/17470218.2015.1022560;

Kuperman, V., & Van Dyke, J. A. (2013). Reassessing word frequency as a determinant of word recognition for skilled and unskilled readers. J*ournal of Experimental Psychology: Human Perception and Performance, 39*(3), 802–823. https://doi.org/10.1037/a0030859;

Laufer, B. & Nation, P. (2001). Passive vocabulary size and speed of meaning recognition. *EUROSLA Yearbook 1,* 7-28;

Laufer, B., & Goldstein, Z. (2004). Testing vocabulary knowledge: Size, strength, and computer adaptiveness. *Language Learning, 54*, 399-436;

Laufer B., & Paribakht, T. S. (1998). The relationship between passive and active vocabularies: Effects of language learning context. *Language Learning, 48*, 365–391;

Lemhöfer, K., & Broersma, M. (2012). Introducing LexTALE: A quick and valid Lexical Test for Advanced Learners of English. *Behavior Research Methods, 44*, 325-343;

Lemhöfer, K., Dijkstra, T., Schriefers, H., Baayen, R. H., Grainger, J., & Zwitserlood, P. (2008). Native language influences on word recognition in a second language: A megastudy. *Journal of Experimental Psychology: Learning, Memory, and Cognition, 34*, 12-31;

Lexical Computing CZ s.r.o. (n.d.) SketchEngine V. 2024. Sketch Engine. Retrieved August 12, 2024, from https://www.sketchengine.eu/;

Ling G., Attali Y., Finn B., Stone E. A. (2017). Is a Computerized Adaptive Test More Motivating Than a Fixed-Item Test? *Applied Psychology Measures 41*(7), 495-511. https://doi.org/10.1177/0146621617707556 ;

Mainz, N., Shao, Z., Brysbaert, M., & Meyer, A. S. (2017). Vocabulary Knowledge Predicts Lexical Processing: Evidence from a Group of Participants with Diverse Educational Backgrounds. *Frontiers in Psychology*, 8, 1164. https://doi.org/10.3389/fpsyg.2017.01164;

Maslennikova E. P., Feklicheva I. W. , E.A. Esipenko E. A., Sharafieva K. R., Ismatullina W. I., Golovin G. W., Mikshalevskiy A. A., Chipeeva N. A., Soldatova E. L. (2017). Slovarnyj zapas kak pokazatel' verbal'nogo intellekta: primenenie ekspress-metodiki ocenki slovarnogo zapasa. [Vocabulary size as a measure of verbal intelligence: application of the express-method of vocabulary size assessment]. *Bulletin of South Ural State University 10*(3), 63–69. Retrieved November 3, 2024, from https://pdfs.semanticscholar.org/45d0/8efe69729db511d98f7d7586bf0bf15ca082.pdf

Masrai, A. (2019). Vocabulary and Reading Comprehension Revisited: Evidence for High-, Mid-, and Low-Frequency Vocabulary Knowledge. *Sage Open, 9*(2). https://doi.org/10.1177/2158244019845182;

Messick, S. (1988), Meaning and values in test validation: the science and ethics of assessment. ETS. Research Report Series, 1988: i-28. https://doi.org/10.1002/j.2330-8516.1988.tb00303.x;

Mizumoto, A., Sasao, Y., & Webb, S. A. (2017). Developing and evaluating a computerized adaptive testing version of the Word Part Levels Test. *Language Testing, 36*(1), 101-123. https://doi.org/10.1177/0265532217725776;

Mochida, K. & Harrington, M. (2006). The Yes/No test as a measure of receptive vocabulary knowledge. *Language Testing, 23*(1), 73-98. https://doi.org/10.1191/0265532206lt321oa;

Monaghan P., Chang Y., Welbourne S., & Brysbaert M. (2017). Exploring the relations between word frequency, language exposure, and bilingualism in a computational model of reading. *Journal of Memory and Language, 93,* 1-21, https://doi.org/10.1016/j.jml.2016.08.003;

Muszyński, M., Banasik-Jemielniak, N., Żółtak, T., Rimfeld, K., Shakeshaft, N. G., Schofield, K. L., Malanchini M., & Pokropek, A. (2023). Moving Intelligence Measurement Online: Adaptation and Validation of the Polish Version of the Pathfinder General Cognitive Ability Test. https://doi.org/10.31234/osf.io/tqyux;

Nation, P. & Beglar, D. (2007). A Vocabulary Size Test. *The Language Teacher, 31*(7), 9-13;

Nation, P. (2024). *Paul Nation's Resources*. Victoria University of Wellington. Retrieved December 2, 2024 from: https://www.wgtn.ac.nz/lals/resources/paul-nations-resources;

Nevin, E., Behan, A., Duffy, G., Farrell, S., Harding, R., Howard, R., Mac Raighne, A., & Bowe, B. (2015). Assessing the validity and reliability of dichotomous test results using Item

Response Theory on a group of first year engineering students. *The 6th Research in Engineering Education Symposium (REES 2015)*, Dublin, Ireland, July 13-15;

Park I. H. (2024). Validation of the Korean Bilingual Version of the Vocabulary Size Test. *English Teaching* 79(2), 139-162. https://doi.org/10.15858/engtea.79.2.202406.139 ;

Parshina O., Ladinskaya N., Gault L., Sekerina I.A. (2024). Predictive Language Processing in Russian Heritage Speakers: Task Effects on Morphosyntactic Prediction in Reading. *Languages,* 9(5):158. https://doi.org/10.3390/languages9050158;

Pellicer-Sánchez, A., & Schmitt, N. (2012). Scoring Yes–No vocabulary tests: Reaction time vs. nonword approaches. *Language Testing,* 29(4), 489-509. https://doi.org/10.1177/0265532212438053

Puig-Mayenco E., Chaouch-Orozco A., Liu H., & Martín-Villenahe M. (2023). The LexTALE as a measure of L2 global proficiency. A cautionary tale based on a partial replication of Lemhöfer and Broersma (2012). *Linguistic Approaches to Bilingualism 13* (3), 299 - 314. https://doi.org/10.1075/lab.22048.pui;

Qi, S., Teng, M. & Fu, A. (2022). LexCH: a quick and reliable receptive vocabulary size test for Chinese Learners. *Applied Linguistics Review.* https://doi.org/10.1515/applirev-2022-0006;

Qian, D. D., & Schedl, M. (2004). Evaluation of an in-depth vocabulary knowledge measure for assessing reading performance. *Language Testing*, 21(1), 28–52. https://doi.org/10.1191/0265532204lt273oa;

Read, J. (2023). Towards a new sophistication in vocabulary assessment. *Language Testing, 40*(1), 40-46. https://doi.org/10.1177/02655322221125698;

Reise, S. P., & Waller, N. G. (2009). Item Response Theory and Clinical Measurement. *Annual Review of Clinical Psychology*, 5(1), 27–48. https://doi.org/10.1146/annurev.clinpsy.032408.153553;

Robitzsch, A., Kiefer, T., & Wu, M. (2020). TAM: Test Analysis Modules. *R package version 3.5-19*. https://CRAN.R-project.org/package=TAM;

Runnels J. (2012). Evaluation of an achievement English vocabulary test using Rasch analysis. *Association of Language and Cultural Education 22,* 155-171. Retrieved September 11,

2024, from http://id.nii.ac.jp/1092/00000937/;

San Mateo-Valdehíta, A. & Criado de Diego, C. (2021). Receptive and productive vocabulary acquisition: effectiveness of three types of tasks. Results from French students of Spanish as secondlanguage. *Onomázein, 51*, 37–56. https://doi.org/10.7764/onomazein.51.05;

Schmitt N. (2014). Size and Depth of Vocabulary Knowledge: What the Research Shows. *Language Learning 64*, 913-951. https://doi.org/10.1111/lang.12077;

Schmitt, N., Schmitt, D., & Clapham, C. (2001). Developing and exploring the behaviour of two new versions of the Vocabulary Levels Test. *Language Testing, 18*(1), 55-88. https://doi.org/10.1177/026553220101800103;

Sohacka K. (2016). A comparison of WISC-R and WAIS-R (PL) scores of children and adolescents in a longitudinal study. *Roczniki psychologiczne/Annals of Psychology 19*(1), 169-178. http://dx.doi.org/10.18290/rpsych.2016.19.1-6pl

Sohacka K. (2019). Problemy z pomiarem inteligencji skalami Wechslera - WISC-R i WAIS-R (Pl). *Eruditio et Ars 1*(2), 48-56. Retrieved August 15, 2024 from http://eruditioetars.ans-ns.edu.pl/images/pdf/2-2019/09_Eruditio_et_Ars_II_Krystyna_SOCHACKA.pdf;

Stoeckel, T., McLean, S., & Nation, P. (2021). Limitations of size and levels tests of written receptive vocabulary knowledge, *Studies in Second Language Acquisition*, *43*(1), 181–203. https://doi.org/10.1017/S027226312000025X;

Straetmans, G. J. J. M., & Eggen, T. J. H. M. (1998). Computerized Adaptive Testing: What It Is and How It Works. *Educational Technology*, *38*(1), 45–52. http://www.jstor.org/stable/44428447;

Tschirner E. (2021). Examining the validity and reliability of ITT vocabulary size tests. Leipzig: Institut für Testforschung und Testentwicklung e.V. Leipzig.

Tseng, W. T. (2016). Measuring English vocabulary size via computerized adaptive testing, *Computers & Educatio 97*, 69-85, https://doi.org/10.1016/j.compedu.2016.02.018 ;

Vermeiren, H., Vandendaele, A. & Brysbaert, M. (2023). Validated tests for language research

with university students whose native language is English: Tests of vocabulary, general knowledge, author recognition, and reading comprehension. *Behavioral Research 55*, 1036–1068. https://doi.org/10.3758/s13428-022-01856-x

Webb, S. (2005). Receptive and Productive Vocabulary Learning: The Effects of Reading and Writing on Word Knowledge. *Studies in Second Language Acquisition, 27*(1), 33–52. https://doi.org/10.1017/S0272263105050023;

Webb, S. (2021). A different perspective on the limitations of size and levels tests of written receptive vocabulary knowledge. *Studies in Second Language Acquisition*, *43*(2), 454–461. https://doi.org/10.1017/S0272263121000449;

Westergaard M. (2024, July 9-10). *Sensitivity to microvariation in gender assignment to ambiguous nouns in Russian: Evidence from monolingual and bilingual native speakers* [Plenary talk]. Psycholinguistic of Slavic Languages, Wrocław, Poland. https://shorturl.at/fqAGY;

Wind S. & Hua C. (2021). *Rasch Measurement Theory Analysis in R: Illustrations and Practical Guidance for Researchers and Practitioners.* https://bookdown.org/chua/new_rasch_demo2/;

Zhao, P., & Ji, X. (2018). Validation of the Mandarin Version of the Vocabulary Size Test. *RELC Journal, 49*(3), 308-321. https://doi.org/10.1177/0033688216639761;

Zhou, C., & Li, X. (2022). LextPT: A reliable and efficient vocabulary size test for L2 Portuguese proficiency. *Behavioral Research 54*, 2625–2639. https://doi.org/10.3758/s13428-021-01731-1